\documentclass[10pt,twocolumn,letterpaper]{article}

\usepackage{cvpr}
\usepackage{times}
\usepackage{epsfig}
\usepackage{graphicx}
\usepackage{amsmath}
\usepackage{amssymb}
\DeclareMathOperator*{\argmax}{argmax}
\usepackage{rotating}
\usepackage{commath}
\usepackage[ruled,vlined]{algorithm2e}
\usepackage{algorithmic}
\usepackage{bbm}
\usepackage{adjustbox}
\usepackage{booktabs}

\usepackage[pagebackref=true,breaklinks=true,letterpaper=true,colorlinks,bookmarks=false]{hyperref}

\cvprfinalcopy 

\def\cvprPaperID{4441} 

\ifcvprfinal\fi

\usepackage{color}

\begin{document}

\title{ Differential Treatment for Stuff and Things: \\ A Simple Unsupervised Domain Adaptation Method for Semantic Segmentation}

\author{Zhonghao Wang$^{1}$, Mo Yu$^{2}$, Yunchao Wei$^{3}$, Rogerio Feris$^{2}$, \\
Jinjun Xiong$^{2}$, Wen-mei Hwu$^{1}$,
Thomas S. Huang$^{1}$, Humphrey Shi$^{4,1}$\\
\\
{\small $^1$C3SR, UIUC, $^2$IBM Research, $^3$ReLER, UTS, $^4$University of Oregon}}

\maketitle
\thispagestyle{empty}

\begin{abstract}
We consider the problem of unsupervised domain adaptation for semantic segmentation by easing the domain shift between the source domain (synthetic data) and the target domain (real data) in this work. State-of-the-art approaches prove that performing semantic-level alignment is helpful in tackling the domain shift issue. 
Based on the observation that stuff categories usually share similar appearances across images of different domains while things (i.e.\ object instances) have much larger differences, we propose to improve the semantic-level alignment with different strategies for stuff regions and for things:
1) for the \textbf{stuff} categories, we generate feature representation for each class and conduct the alignment operation from the target domain to the source domain; 2) for the \textbf{thing} categories, we generate feature representation for each individual instance and encourage the instance in the target domain to align with the most similar one in the source domain. In this way, the individual differences within thing categories will also be considered to alleviate over-alignment. In addition to our proposed method, we further reveal the reason why the current adversarial loss is often unstable in minimizing the distribution discrepancy and show that our method can help ease this issue by minimizing the most similar stuff and instance features between the source and the target domains. We conduct extensive experiments in two unsupervised domain adaptation tasks, \ie GTA5 $\rightarrow$ Cityscapes and SYNTHIA $\rightarrow$ Cityscapes, and achieve the new state-of-the-art segmentation accuracy. 
Our code will be avaiable at \href{https://github.com/SHI-Labs/Unsupervised-Domain-Adaptation-with-Differential-Treatment}{https://github.com/SHI-Labs/Unsupervised-Domain-Adaptation-with-Differential-Treatment}.

\end{abstract}
\vspace{-4mm}
\begin{figure}
\centering
\includegraphics[width=0.45\textwidth]{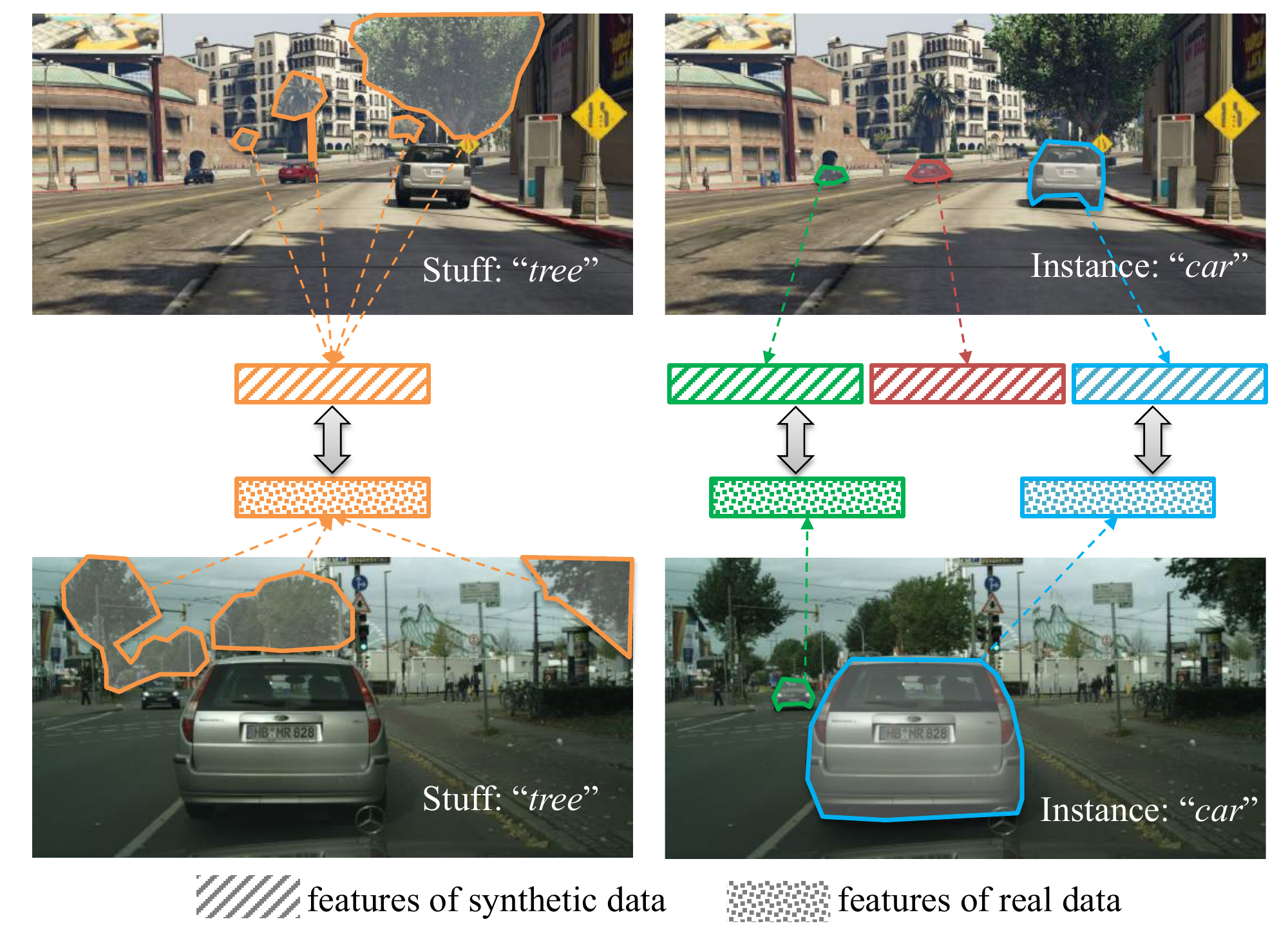}
\caption{Illustration of the proposed Stuff Instance Matching (SIM) structure. By matching the most similar stuff regions and things (i.e., instances) with differential treatment, we can adapt the features more accurately from the source domain to the target domain. }
\label{teaser}
\vspace{-3mm}
\end{figure}

\section{Introduction}
Semantic segmentation \cite{Long_2015_CVPR} enables image scene understanding at the pixel level, which is crucial to many real-world applications such as autonomous driving. The recent surge of deep learning \cite{deeplearning} methods that generate features from large training datasets has significantly accelerated the progress in semantic segmentation \cite{deeplabv1, deeplabv2, pspnet,deeplabv3,huang2019ccnet,huang2020alignseg,cheng2019spgnet,wei2018revisiting,2019arXiv191110194C,jiao2019geometry,qian2019weakly}. However, collecting data with pixel-level annotation is costly in terms of both time and money. Specifically, to annotate an image in the widely used benchmark Cityscapes \cite{cityscapes} takes 1.5 hours on average; that sums up to 7,500 hours in total for annotating all the 5,000 images. 
Such annotation cost is quite burdensome, given that training deep neural networks on the collected data usually takes less than dozens of hours.

To address the problem of high-cost annotation, unsupervised domain adaptation methods are proposed for semantic segmentation~\cite{gta, synthia}.
In these works, a model trained on a source domain dataset with segmentation annotations is adapted for an unlabeled target domain. The source domain datasets can be synthetic, e.g., from video games, so that little human effort is required.
However, such methods suffer from the domain shift problem. Existing methods deal with the problem by minimizing the distribution discrepancy of the features extracted by a feature extractor \cite{vgg,resnet} between the source domain and the target domain. To this end, the GAN \cite{gan} architectures, usually composed of a generator and a discriminator, are broadly used in this context.
The generator extracts features from the input images, and the discriminator distinguishes which domain the features are generated from. 
The discriminator can thereby guide the generator to generate the target domain features with a distribution closer to the feature distribution of the source domain 
in an adversarial way.

\begin{figure}
\centering
\includegraphics[width=0.45\textwidth]{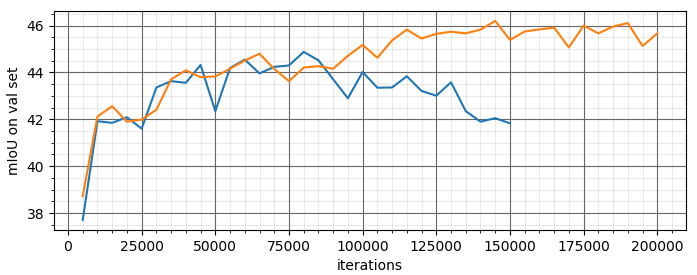}
\caption{mIoU comparison on the validation set of Cityscapes by adapting from GTA5 dataset to Cityscapes dataset. The blue line corresponds to the output space adversarial adaptation strategy \cite{outputspace}. The orange line corresponds to the output space adversarial adaptation combined with our proposed SIM structure. 
The model performance is tested every 5000 iterations.}
\label{miou_curve}
\vspace{-4mm}
\end{figure}

In the previous GAN-style approaches, the adversarial loss is essentially a binary cross-entropy about whether the generated feature is from the source domain.
We observe that such a global training signal is usually weak for the segmentation task. 
First, the alignments between stuff regions and between things require different treatments but the adversarial loss lacks such structural information.
For example, the stuff regions usually lack the appearance variance in an image but the things can have diverse appearances in the same image. 
Therefore, it is sub-optimal to use an adversarial loss to align the stuff and thing features globally without differential treatments. 
%
Second, 
the global GAN structure only adapts the feature distribution between two domains and does not necessarily adapt the target domain features towards the most likely space of source domain features. Therefore, as the semantic head gathers the features from the source domain with more training iterations, it becomes harder for the feature generator to adapt the target domain features exactly toward the source domain features. This leads to a performance drop on the target domain images as shown in figure \ref{miou_curve}.  

This paper proposes a stuff and instance matching (SIM) framework to address the aforementioned difficulties. First, we treat the alignments between stuff regions and between instances of things with different guidance. The key idea is shown in figure \ref{teaser}. The multiple stuff regions in a source image are usually similar, so the stuff from different domains can be directly aligned with their global feature vectors. While the multiple instances of the same thing, e.g., of the car category, can be diverse in the source image. Therefore we align instances in the target image to the most similar ones in the source image.

Second, we deal with the instability with the GAN training framework, we apply a L1 loss to explicitly minimize the distance between the target domain stuff and thing features with the most similar source domain counterparts. In this way, the adaptation is processed in a more accurate direction, instead of the rough distribution matching when using only the adversarial cross entropy loss, even after the semantic head gathers the source domain features with longer training iterations. As shown in figure \ref{miou_curve}, we implement the output space adversarial adaptation \cite{outputspace} from GTA5 \cite{gta} dataset to Cityscapes \cite{cityscapes} dataset, and compare it with our model which adds the SIM module. We successfully solve the problem of the performance drop at longer training iterations with few more computations.

Finally, we propose to improve the SIM framework with a self-supervised learning strategy. Specifically, we use predicted segmentation with high confidence to train the segmentation model, and to enhance the alignment for both stuff categories and thing categories.

We evaluate the proposed approach on two unsupervised domain adaptation tasks, the adaptation from GTA5 to Cityscapes and from SYNTHIA to Cityscapes, and achieve a new state-of-the-art performance on both tasks.

\begin{figure*}
\centering

\includegraphics[width=1\textwidth]{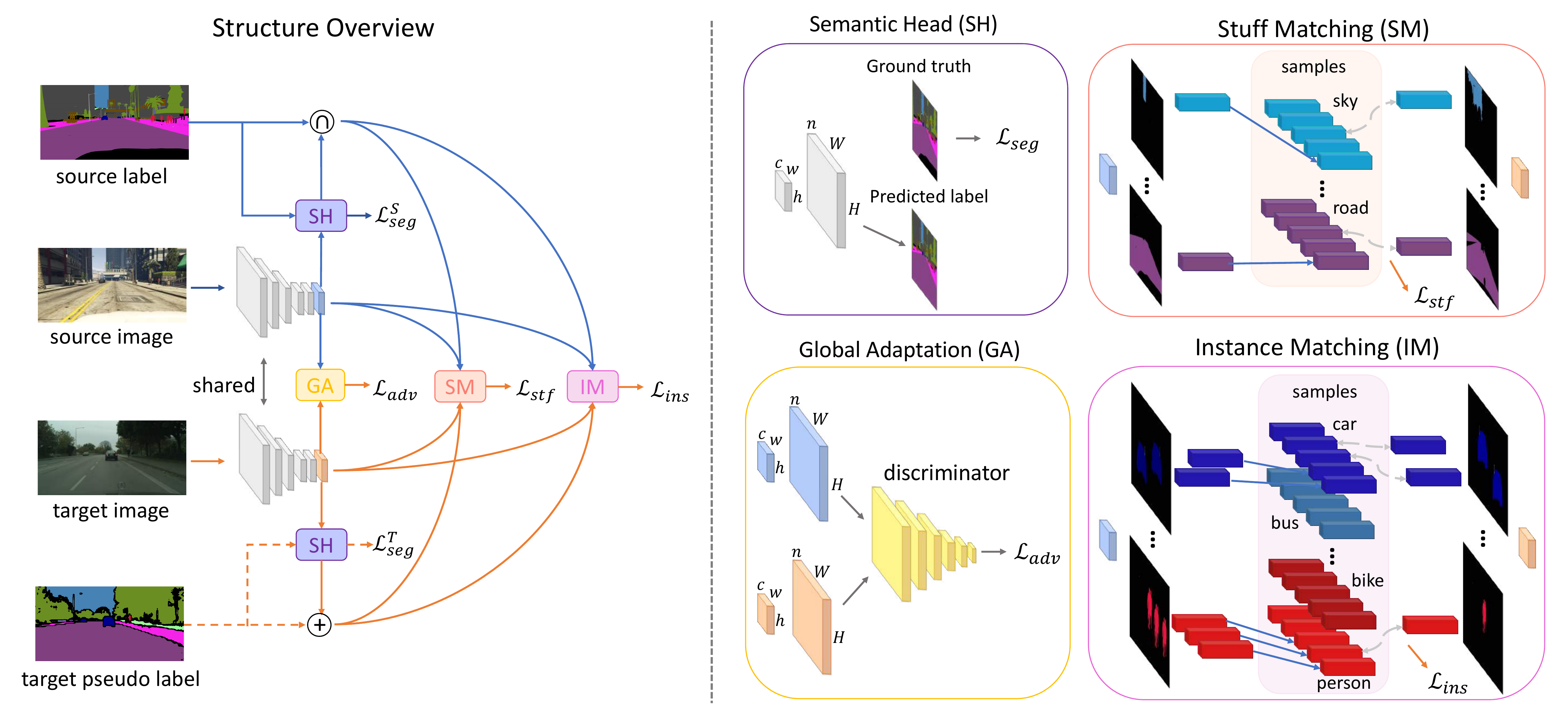}
\caption{Framework. 1) The overall structure is shown on the left. The solid lines represent the first step training procedure in Eqn (\ref{Lstep1}), and the dash lines along with the solid lines represent the second step training procedure in Eqn (\ref{Lstep2}). The blue lines correspond to the flow direction of the source domain data, and the orange lines correspond to the flow direction of target domain data. $\cap$ is an operation defined in Eqn (\ref{matchgp}); $+$ is an operation defined in Eqn (\ref{labelaug}) and is only effective in the second step training procedure. 2) The specific module design is shown on the right. $h$, $w$ and $c$ represent the height, width and channels for the feature maps; $H$, $W$ and $n$ represent the height, width and class number for the output maps of the semantic head. For SH, the input ground truth label map supervise the the semantic segmentation task, and the semantic head also generates a predicted label map joining the operations of $\cap$ and $+$. For SM and IM, the grey dash lines represent the matching operation defined in Eqn (\ref{Lcls}) and (\ref{Lins}) respectively.}
\label{frameworks}
\vspace{-3mm}
\end{figure*}
\section{Related works}
\label{relatedworks}
The domain adaptation in classification is a broadly studied problem after the surge of deep learning methods and a big progress has been made \cite{clsadaptsurvey}. However, the domain adaptation in semantic segmentation problem is more challenging as it is in essence a pixel-level classification problem involving structured contextual semantic adaptation. A typical practice of this task is adapting a semantic segmentation model trained on synthetic datasets \cite{gta,synthia} (source domain) to perform on real image datasets \cite{cityscapes} (target domain). The key idea of the domain adaptation task is to align the feature distributions between the source domain and the target domain, so that the model can utilize the knowledge learned from the source domain to perform tasks on the target domain. We generally divide current methods into three categories: image-level transferring, feature-level transferring and label-level transferring.

The image-level transferring refers to changing the appearance of images such that images from the source domain and the target domain are more visually similar. These methods \cite{bidir, dcan, aan} usually transfer the color, illumination and other stylization factors of images from one domain to another or from both domains to a neutral domain. In \cite{bidir}, Li et al. use CycleGAN \cite{cyclegan} with a perceptual loss to preserve the locality of semantic information to perform the unpaired image-to-image transferring. In \cite{aan}, Zhang et al. propose an Appearance Adaptation Network which transfers appearances of images between two domains mutually, such that the images appearance tend to be domain-invariant. Choi et al.~\cite{selfensemble} raise a GAN-based self-ensembling data augmentation method for domain alignment. 

The feature-level transferring refers to matching the extracted feature distributions between the source domain and the target domain. While feature extractors \cite{vgg, resnet, densenet} can extract task-specific features, the features extracted from the target domain and the ones from the source domain have a discrepancy due to the domain shift, which negatively impacts the model's performance on the target domain dataset. Therefore, minimizing the feature distribution discrepancy with GAN \cite{gan} structure is a common practice in domain adaptation. Sankaranarayanan et al. proposes an image reconstruction framework \cite{Sankaranarayanan} to make the reconstructed images from two domains close to each other so that the features are pulled closer with back propagation. Tsai and et al. proposes a simple end-to-end output space domain adaptation framework \cite{outputspace}. Wu and et al. proposes a channel-wise feature alignment network \cite{dcan} to close the gap of the channel-wise mean and standard deviation in CNN feature maps. Chang and et al. propose a framework \cite{allaboutstructure} to extract domain-invariant structures for adaptation. 

The label-level transferring refers to giving pseudo-labels to the target domain dataset given the knowledge learned from the source domain for helping the adaptation task. This follows a self-supervised learning framework \cite{selfsurvey} where no human efforts are input for labeling the target dataset. Zou et al. \cite{Zou_2018_ECCV} proposes a class-balanced self-training framework. Li et al. \cite{bidir} proposes a joint self-learning and image transferring frameworkfor adaptation. 

\vspace{-3mm}

\section{Background}
\label{section31}

\paragraph{Definitions}
We follow the unsupervised semantic segmentation framework for the domain adaptation task; that is, given a source domain dataset with images and the pixel-level semantic annotations $\{x^s_i, y^s_i\}$ and a target domain dataset with only images $\{x^t_i\}$, we plan to train a model that can predict the pixel-level labels $\{\hat{y}^t_i\}$ for the target domain images. We denote the class number with $N$. 
\vspace{-4mm}
\paragraph{Segmentation and adversarial adaptation}
The semantic segmentation task in deep learning literature is broadly discussed \cite{deeplabv1, deeplabv2, pspnet, deeplabv3}, and the problem solving strategy is formalized by utilizing a feature extractor network $F$ to extract image features and a classification head $C$ to classify features into semantic classes. We use the cross entropy loss to supervise the model on the pixel classification task with the annotated source domain dataset in Eqn (\ref{Lseg}). 
\begin{equation}
    \mathfrak{L}_{seg}^S(f^s_i) = -\sum_{i,h,w} \sum_{k\in N} y^{(h,w)}_i\log (\mathcal{S}(C(f^{s}_i)^{(h,w)})^{(k)})
    \label{Lseg}
    \vspace{-1mm}
\end{equation}
where $f_i^s = F(x_i^s)$, $x_i^s\in X^s$, $X^s$ is the source domain image dataset, $h$ and $w$ are the height and width of the feature maps, $y$ is the ground truth label, $\mathcal{S}$ is the softmax operation. However, due to the domain shift problem, the model trained on the source domain will achieve inferior performance if directly applied to test on the target domain. Therefore, we impose a traditional GAN structure on the output space \cite{outputspace} to globally minimize the feature distribution discrepancy between the source domain and the target domain. Here, the feature extractor $F$ and the classification head $C$ serve as the generator $G$ where $G=C\circ F$. A discriminator $D$ will discriminate the generated output by the generator $G$. We close the feature distribution discrepancy between the source domain and the target domain by optimizing the adversarial target function in Eqn (\ref{Ladv}).
\vspace{-2mm}
\begin{equation}
    \min_{G} \mathfrak{L}_{adv}(G, D) = -\sum_{x_i^t\in X^T} \log (1-D(\mathcal{S}(G(x_i^t))))
    \label{Ladv}
    \vspace{-2mm}
\end{equation}
while the discriminator tries to distinguish which domain the feature is from by optimizing the discriminator target function in Eqn (\ref{Ld}).
\vspace{-2mm}
\begin{equation}
\begin{split}
    \min_{D} \mathfrak{L}_{D}(G, D) = -\sum_{x_i^t\in X^T} \log (D(\mathcal{S}(G(x_i^t)))) \\
    - \sum_{x_j^t\in X^S} \log (1-D(\mathcal{S}(G(x_j^s)))) 
    \label{Ld}
\end{split}
\vspace{-3mm}
\end{equation}

\section{Proposed Methods}

The key idea of our method is that the past experience leading to good outcomes should also help the current training process. Specifically to our task, the past experience should help both the feature-level transferring and the label-level transferring from the source domain to the target domain. First, we raise a stuff and instance matching (SIM) framework to reduce the intra-class domain shift problem. Second, we propose a self-supervised learning framework combined with our proposed SIM structure to enable the label-level transferring, which further boosts the performance. The overall framework is shown in figure \ref{frameworks}.

\subsection{Stuff and instance matching (SIM)}
\label{section32}

First, we discuss the matching process for the background classes such as road, sidewalk, sky and etc.. These classes usually cover a large area of the image and lack appearance variation, so we only extract the image-level stuff feature representation for them. For each source domain image, we access the correctly classified label map by selecting the predicted labels matched with the ground truth labels in Eqn (\ref{matchgp}). 
\vspace{-2mm}
\begin{equation}
\begin{split}
    L_{P_i}^{s} &= \argmax_{k\in N} (C(f_i^{s})^{(k)}) \\
    L_{C_i}^{s} &= L_{G_i}^s \cap L_{P_i}^s
    \label{matchgp}
\end{split}
\vspace{-2mm}
\end{equation}
where $L_{C_i}^s$ is the correctly classified label map, $L_{G_i}^s$ is the ground truth label map, $L_{P_i}^s$ is the predicted label map, and $i\in \{1..|X^S|\}$. We average the features belonging to the same background semantic class across the width and height of the image as the stuff representation for each background class in Eqn (\ref{semanticave}).
\vspace{-2mm}
\begin{equation}
\begin{split}
    \mathcal{A}^b(L,f)  &= \frac{\sum_{h,w}\delta({L}^{(h,w)}-b){f}^{(h,w)}}{\max(\epsilon, \sum_{h,w}\delta({L}^{(h,w)}-b))}\\
    S^b_{j} &= \mathcal{A}^b(L_{C_i}^s,f_i^s) \;\; \textbf{where}\; j=i\, \textrm{mod}\, w, \\
    &\quad\quad\quad\quad\quad\quad\quad\textbf{if}\;  \mathcal{A}^b(L_{C_i}^s,f_i^s)\neq0
\end{split}
\label{semanticave}
\vspace{-2mm}
\end{equation}
where $S^b_j$ is the $j$'th source domain semantic feature sample of class $b$, $b \in B$ (background classes), $i\in\{1..|X^S|\}$, $w$ is the number of feature samples to be stored for each class, $\delta$ is the Dirac delta function and $\epsilon$ is a regularizing term. For each target domain image, we minimize the distance of the stuff representation of each background class with the closest intra-class source stuff feature representation. Because the ground truth of the target domain image is not provided, we use the predicted label map to generate the stuff feature representation for each background class. We adapt the stuff feature representation of the background classes by minimizing the loss function defined in Eqn (\ref{Lcls}) when the model is trained on the target domain.
\vspace{-2mm}
\begin{equation}
    \mathfrak{L}_{stf} = \sum_{i}\sum_{b}\min_{j}\norm{\mathcal{A}^b(L_{P_i}^t,f_i^t)-S^b_{j}}^1_1
    \label{Lcls}
    \vspace{-2mm}
\end{equation}
where $i\in \{1..|X^T|\}$, and $b\in L_{P_i}^t \cap B$.

Second, we discuss the instance matching process for the foreground classes such as cars, persons and etc.. Because the ground truth does not provide the instance level annotations, we generate the foreground instance mask by finding the disconnected regions for each foreground class in the label map $L$. This coarsely segment the intra-class semantic regions into multiple instances, and thus various instance-level feature representations of one image can be generated accordingly in Eqn (\ref{regionfeature}).
\vspace{-2mm}
\begin{equation}
\begin{split}
    R_k = \{r_{k_1}, r_{k_2}, ..., r_{k_m}\} = \mathcal{T}(L, k) \\ 
    \mathcal{I}(r,f) = \frac{\sum_{h,w} r^{(h,w)}f^{(h,w)}}{\max({\epsilon, \sum_{h,w} r^{(h,w)}})}
\end{split}
\label{regionfeature}
\vspace{-2mm}
\end{equation}
where $r_{k_i}$ is the $i$'th ($i\in\{1,..,m\}$) binary mask of the connected region belonging to class $k$, $k\in K$ (foreground classes), $\mathcal{T}$ is the operation to find the disconnected regions of class $k$ from the label mask $L$, and $\mathcal{I}$ is the operation to generate the instance-level feature representation. The source domain instance feature samples can be generated in algorithm \ref{alg1}.
Therefore, the target domain instance features can be pulled closer to the closest intra-class source domain instance feature sample by minimizing the loss function in Eqn (\ref{Lins}).
\vspace{-2mm}
\begin{equation}
    \mathfrak{L}_{ins} = \sum_{i}\sum_{k\in K}\frac{1}{\abs{R^t_k}}\sum_{r^t\in R^t_k}\min_{j}\norm{\mathcal{I}(r^t,f_i^t)-S^k_{j}}^1_1
    \label{Lins}
    \vspace{-2mm}
\end{equation}
where $i\in \{1..|X^T|\}$, and $R_k^t=\mathcal{T}(L_{P_i}^t, k)$.

\begin{algorithm}[]
\SetAlgoLined
\KwResult{$S^k$}
 $z = 10$; \emph{\# maximum class instances in an image} \\
 $c_k = 0, \forall k\in K$; \emph{\# instance feature counter}\\
 \For {$x_i^s \in X^S$}{
  \For {$k \in K$}{
    $R_k^s = \mathcal{T}(L_{C_i}^s, k)$\\
    \If{$R_k^s \neq \emptyset$}{
      $R_{sort}$ = sort $R_k^s$ by area in descent order\\
      \For {$l \in \{1..\min(z, \abs{R_{sort}})\}$}{
        $j = c_k \mod z*w$\\
        $c_k = c_k+1$\\
        $S^k_{j} = \mathcal{I}(R_{sort}[l],f_i^s)$ \\
      }
    }
  }
 }
 \caption{Instance-level source feature samples}
 \label{alg1}
\end{algorithm}

\subsection{Self-supervised learning with SIM}
Because the model is only trained on the source domain with the ground truth annotations, the features and the softmax output are thus generated to optimize the source domain segmentation loss function but ignore the target domain segmentation supervision. However, the distribution of the ground truth labels from both domains also have a discrepancy, and this negatively impacts the model's performance on the target domain. Therefore, we propose a self supervised learning framework combined with our feature matching methods to alleviate this problem. 

We first follow the framework described in sections \ref{section31} and \ref{section32} to train a model with the source domain images $X^S$ and ground truth annotations $Y^S$ along with the target domain images $X^T$. Then we use the trained model to give pseudo-labels to the pixels with high confidence of the predicted labels in the training set images $X^T$ shown in Eqn (\ref{labelpseudo}). 
\vspace{-2mm}
\begin{equation}
    \hat{y}^t_i = \argmax_{k\in N} \mathbbm{1}_{[\mathcal{S}(C(f_i^{t}))^{(k)}> y_t^k]}(C(f_i^{t})^{(k)})
    \label{labelpseudo}
    \vspace{-2mm}
\end{equation}
where $\mathbbm{1}$ is a function which returns the input if the condition is true or a don't care symbol if not, and $y_t^k$ is the confidence threshold for class $k$. Then, we add the semantic segmentation loss on the target domain images in Eqn (\ref{Lpseudo}) along with other losses to retrain our model.
\vspace{-2mm}
\begin{equation}
    \mathfrak{L}_{seg}^T(f^t) = -\sum_{i,h,w} \sum_{k\in N} \hat{y}^{(h,w)}_i\log (\mathcal{S}(C(f^{t}_i)^{(h,w)})^{(k)})
    \label{Lpseudo}
    \vspace{-2mm}
\end{equation}
With the pseudo labels supervising the model to generate features corresponding to specific classes, these features should generically be adapted to be closer to the corresponding intra-class source domain features. The $L_{P_i}^t$ is thereby augmented by Eqn (\ref{labelaug}) for the stuff feature adaptation loss defined in Eqn (\ref{Lcls}) and the instance feature adaptation loss defined in Eqn (\ref{Lins}):
\vspace{-2mm}
\begin{equation}
    \mathbbm{1}_{L_{P_i}^t\neq \hat{y}_i^t}(L_{P_i}^t) = \mathbbm{1}_{L_{P_i}^t\neq \hat{y}_i^t}(\hat{y}_i^t).
    \label{labelaug}
    \vspace{-2mm}
\end{equation}
$\mathbbm{1}$ selects the positions in the input satisfying the condition.

\subsection{Training procedure}
We follow a two-step training procedure to improve the performance of the generator $G$ on semantic segmentation task on the target domain dataset. First, we train our model without the self-supervised learning module, and optimize the target function in Eqn (\ref{Lstep1}) with $G$ and $D$ in an adversarial training strategy:
\vspace{-2mm}
\begin{equation}
\begin{split}
    \min_{G,D}\mathfrak{L}_{step1} = &\min_{G}(\lambda_{seg}\mathfrak{L}_{seg}^S + \lambda_{adv}\mathfrak{L}_{adv} + \\
    &\lambda_{ci}(\mathfrak{L}_{stf} + \mathfrak{L}_{ins})) + \min_{D} \lambda_{D}\mathfrak{L}_{D},
\end{split}
    \label{Lstep1}
    \vspace{-2mm}
\end{equation}
where $\lambda$'s are the weight parameters for the losses. Second, after giving the pseudo labels to the target domain training dataset with the model trained in the first step, we reinitialize and repeat the training process to optimize the loss function in Eqn (\ref{Lstep2}).
\vspace{-2mm}
\begin{equation}
\begin{split}
    \min_{G,D}\mathfrak{L}_{step2} = &\min_{G}(\lambda_{seg}(\mathfrak{L}_{seg}^S+\mathfrak{L}_{seg}^T) + \lambda_{adv}\mathfrak{L}_{adv} + \\
    &\lambda_{ci}(\tilde{\mathfrak{L}}_{stf} + \tilde{\mathfrak{L}}_{ins})) + \min_{D} \lambda_{D}\mathfrak{L}_{D} ,
\end{split} \label{Lstep2}
\vspace{-2mm}
\end{equation}
where $\tilde{\mathfrak{L}}_{stf}$ and $\tilde{\mathfrak{L}}_{ins}$ are augmented with predicted $\hat{y}^t_i$s according to Eqn (\ref{labelaug}).

\begin{table*}[t!]
\caption{Comparison to the state-of-the-art results of adapting GTA5 to Cityscapes. }
\small
\begin{center}
\begin{tabular}{ l|p{0.28cm}p{0.28cm}p{0.28cm}p{0.28cm}p{0.28cm}p{0.28cm}p{0.28cm}p{0.28cm}p{0.28cm}p{0.28cm}p{0.28cm}p{0.28cm}p{0.28cm}p{0.28cm}p{0.28cm}p{0.28cm}p{0.28cm}p{0.28cm}p{0.28cm}c }
 \toprule
 \multicolumn{21}{c}{GTA5 $\rightarrow$ Cityscapes} \\
 \midrule
Method & \begin{turn}{45}road\end{turn}
& \begin{turn}{45}sidewalk\end{turn}& \begin{turn}{45}building\end{turn}& \begin{turn}{45}wall\end{turn}& \begin{turn}{45}fence\end{turn}& \begin{turn}{45}pole\end{turn}& \begin{turn}{45}light\end{turn}& \begin{turn}{45}sign\end{turn}& \begin{turn}{45}vegetation\end{turn}& \begin{turn}{45}terrain\end{turn}& \begin{turn}{45}sky\end{turn}& \begin{turn}{45}person\end{turn}& \begin{turn}{45}rider\end{turn}& \begin{turn}{45}car\end{turn}& \begin{turn}{45}truck\end{turn}& \begin{turn}{45}bus\end{turn}& \begin{turn}{45}train\end{turn}& \begin{turn}{45}motorbike\end{turn}& \begin{turn}{45}bike\end{turn}& mIoU
\\
\midrule
Wu et al.\cite{wu}&85.0&30.8&81.3&25.8&21.2&22.2&25.4&26.6&83.4&36.7&76.2&58.9&24.9&80.7&29.5&42.9&2.5&26.9&11.6&41.7
\\
Tsai et al.\cite{outputspace}&86.5&36.0&79.9&23.4&23.3&23.9&35.2&14.8&83.4&33.3&75.6&58.5&27.6&73.7&32.5&35.4&3.9&30.1&28.1&42.4
\\
Saleh et al.\cite{saleh}&79.8&29.3&77.8&24.2&21.6&6.9&23.5&44.2&80.5&38.0&76.2&52.7&22.2&83.0&32.3&41.3&\textbf{27.0}&19.3&27.7&42.5
\\
Luo et al. \cite{Luo_2019_CVPR} &88.5& 35.4& 79.5& 26.3& 24.3& 28.5& 32.5& 18.3& 81.2& 40.0& 76.5& 58.1& 25.8& 82.6& 30.3& 34.4& 3.4& 21.6& 21.5& 42.6
\\
Hong et al.\cite{hong}&89.2&\textbf{49.0}&70.7&13.5&10.9&38.5&29.4&33.7&77.9&37.6&65.8&\textbf{75.1}&32.4&77.8&39.2&45.2&0.0&25.5&35.4&44.5
\\
Chang et al. \cite{allaboutstructure}&\textbf{91.5}& 47.5& 82.5& 31.3& 25.6& 33.0& 33.7& 25.8& 82.7& 28.8& 82.7& 62.4& 30.8& 85.2& 27.7& 34.5& 6.4& 25.2& 24.4& 45.4
\\
Du et al. \cite{Du_2019_ICCV}& 90.3& 38.9& 81.7& 24.8& 22.9& 30.5& 37.0& 21.2& 84.8& 38.8& 76.9& 58.8& 30.7& 85.7& 30.6& 38.1& 5.9& 28.3& 36.9& 45.4
\\
Vu et al. \cite{Vu_2019_CVPR}& 89.4& 33.1& 81.0& 26.6& 26.8& 27.2& 33.5& 24.7& 83.9& 36.7& 78.8& 58.7& 30.5& 84.8& 38.5& 44.5& 1.7& 31.6& 32.4& 45.5
\\
Chen et al. \cite{Chen_2019_ICCV}& 89.4& 43.0& 82.1& 30.5& 21.3& 30.3& 34.7& 24.0& 85.3& 39.4& 78.2& 63.0& 22.9& 84.6& 36.4& 43.0& 5.5& \textbf{34.7}& 33.5& 46.4
\\
Zou et al. \cite{Zou_2018_ECCV}&89.6& 58.9& 78.5& 33.0& 22.3& \textbf{41.4}& \textbf{48.2}& \textbf{39.2}& 83.6& 24.3& 65.4& 49.3& 20.2& 83.3& 39.0& 48.6& 12.5& 20.3& 35.3& 47.0
\\
Lian et al. \cite{Lian_2019_ICCV} &90.5& 36.3& 84.4& 32.4& \textbf{28.7}& 34.6& 36.4& 31.5& \textbf{86.8}& 37.9& 78.5& 62.3& 21.5& \textbf{85.6}& 27.9& 34.8& 18.0& 22.9& \textbf{49.3}& 47.4
\\
Li et al. \cite{bidir} &91.0& 44.7& 84.2& \textbf{34.6}& 27.6& 30.2& 36.0& 36.0& 85.0& \textbf{43.6}& 83.0& 58.6& \textbf{31.6}& 83.3& 35.3& \textbf{49.7}& 3.3& 28.8& 35.6& 48.5
\\
\hline
ours (ResNet101)&90.6&44.7&\textbf{84.8}&34.3&\textbf{28.7}&31.6&35.0&37.6&84.7&43.3&\textbf{85.3}&57.0&31.5&83.8&\textbf{42.6}&48.5&1.9&30.4&39.0&\textbf{49.2}\\
\midrule
Du et al. \cite{Du_2019_ICCV} & 88.7 & 32.1 & 79.5 & 29.9 & 22.0 & 23.8 & 21.7 & 10.7 & 80.8 & 29.8 & 72.5 & 49.5 & 16.1 & 82.1 & 23.2 & 18.1 & 3.5 & 24.4 & 8.1 & 37.7 \\
Li et al. \cite{bidir} & 89.2 & 40.9 & 81.2 & 29.1 & 19.2 & 14.2 & 29.0 & 19.6 & 83.7 & 35.9 & 80.7 & 54.7 & 23.3 & 82.7 & 25.8 & 28.0 & 2.3 & 25.7 & 19.9 & 41.3 \\
\hline
ours (VGG16) & 88.1 & 35.8 & 83.1 & 25.8 & 23.9 & 29.2 & 28.8 & 28.6 & 83.0 & 36.7 & 82.3 & 53.7 & 22.8 & 82.3 & 26.4 & 38.6 & 0.0 & 19.6 &17.1 & 42.4\\
\bottomrule

\end{tabular}
\end{center}
\label{gta2city}
\vspace{-6mm}
\end{table*}

\section{Implementation}
\subsection{Network architecture}
\textbf{Segmentation Network.} We adopt ResNet-101 model \cite{resnet} pre-trained on ImageNet \cite{imagenet} with only the 5 convolutional layers \{$conv1$, $res2$, $res3$, $res4$, $res5$\} as the backbone network. Due to memory limit, we do not use the multi-scale fusion strategy \cite{multiscale}. For generating better-quality feature maps, we follow the common practice from \cite{deeplabv1, multiscale, outputspace} and twice the resolution of the feature maps of the final two layers. To enlarge the field of view, we use dilated convolutional layers \cite{multiscale} with stride 2 and 4 in $res4$ and $res5$. For the classification heads, we apply an ASPP module \cite{deeplabv2} to $res5$ with $\lambda_{seg}=1$. 

\textbf{Discriminator.} Following \cite{outputspace}, We use 5 convolutional layers with kernel size 4$\times$4, stride of 2 and channel number of \{64, 128, 256, 512, 1\} respectively to form the network. We use a leaky ReLU \cite{leakyrelu} layer of 0.2 negative slope between adjacent convolutional layers. Due to the small batch size in the training process, we do not use batch normalization layers \cite{batch}. The sole discriminator is implemented on the upsampled softmax output of the ASPP head on $res5$ with $\lambda_{adv}=0.001$ and $\lambda_{D}=1$. 

\subsection{Training Details}
We use Pytorch toolbox and a single GPU to train our network. Stochastic Gradient Descent (SGD) is used to optimize the segmentation network. We use Nesterov's method \cite{Nesterov} with momentum 0.9 and weight decay $5\times10^{-4}$ to accelerate the convergence. Following \cite{deeplabv1}, we set the initial learning rate to be $2.5\times 10^{-4}$ and let it polynomially decay with the power of 0.9. For the discriminator networks, we use Adam optimizer \cite{adam} with momentum 0.9 and 0.99. The initial learning rate is set to $10^{-4}$ and the same polynomial decay rule is applied.

\section{Experiments}
\subsection{Datasets}
The Cityscapes \cite{cityscapes} dataset consists of 5000 images of resolution $2048\times1024$ with high-quality pixel-level annotations. These images of street scenes were annotated with 19 semantic labels for evaluation. This dataset is split into training, validation and test sets with 2975, 500 and 1525 images respectively. Following previous works \cite{fcnwild, domainchallenge}, We only evaluate our models on the validation set. The GTA5 \cite{gta} dataset contains 24966 fine annotated synthetic images of resolution $1914\times1052$. All the images are frames captured from the game Grand Theft Auto V. To accommodate the model with the limited GPU memory, we follow \cite{outputspace} and resize GTA5 images to the resolution of $1280\times 720$. This dataset shares all the 19 classes used for evaluation in common with the Cityscapes dataset. The SYNTHIA \cite{synthia} dataset has 9400 images of resolution $1280\times760$ with pixel-level annotations. Similar to \cite{Luo_2019_CVPR, outputspace,Du_2019_ICCV,bidir}, we evaluate our models on Cityscapes validation set with the 13 classes shared in common between SYNTHIA dataset and Cityscapes dataset. The Cityscapes images are resized to $1024\times 512$ for both the training stage and the testing stage.

\subsection{GTA5 to Cityscapes}
\label{section52}
We first show our over results and compare to the previous state-of-the-arts; then discuss the effectiveness of each module in our model; finally we discuss the choice of hyper parameters of our proposed SIM module.

\textbf{Overall results}. We compare the performance of our method with the current state-of-the-arts in table \ref{gta2city}. For fair comparison, we list the performance of the models using resnet-101 \cite{resnet} and VGG16 \cite{vgg} as the backbones respectively. Our method achieves the state-of-the-art performance with either backbone. 

\textbf{Module contributions}. We show the contribution of each module to the overall performance of our model in table \ref{ablation}. If trained purely on the source domain dataset, the model can achieve an mIoU of 36.6 on the Cityscapes validation set. Then, we follow the work of \cite{outputspace} to add the global adversarial training on the output space with the adversarial loss in Eqn (\ref{Ladv}) and the discriminator loss in Eqn (\ref{Ld}), and the mIoU is thereby improved to 41.4. As mentioned in section \ref{relatedworks}, image-level adaptation is also a key factor in minimizing the discrepancy of data distribution. Therefore, it is helpful to utilize a transferred source-domain image dataset whose appearance is more similar to that of the target-domain image dataset. We adopt the transferred GTA5 images of \cite{bidir} which utilizes a CycleGAN \cite{cyclegan} structure to adapt the style of GTA5 images to the style of Cityscapes images. This further improves the mIoU to 44.9, which serves as the baseline for our works. 

\begin{table}[t!]
    \centering
    \caption{Ablation study on the adaptation from GTA5 dataset to Cityscapes dataset. AA stands for adversarial adaptation; IT stands for image transferring; SIM stands for semantic and instance matching; SSL stands for self-supervised learning.}
    \small
    \begin{tabular}{l|cccc|c}
        \toprule
         method & AA & IT & SIM & SSL & mIoU  \\
         \midrule
         source only&&&&&36.6\\
         \,+ AA\cite{outputspace}&\checkmark&&&&41.4\\
         \,+ IT\cite{bidir}&\checkmark&\checkmark&&&44.9\\
         \midrule
         \,+ SIM&\checkmark&\checkmark&\checkmark&&46.2\\
         \,+ SSL&\checkmark&\checkmark&\checkmark&\checkmark&49.2\\
         \midrule
         target only&&&&&65.1\\
         \bottomrule
    \end{tabular}
    \label{ablation}
    \vspace{-3mm}
\end{table}
\begin{table}[t!]

    \centering
    \caption{Influence of $\lambda_{ci}$ given the number of semantic feature samples to be stored is 50 ($w=50$)}
    \small
    \begin{tabular}{l|ccccc}
        \toprule
         $\lambda_{ci}$&0.1&0.05&0.01&0.005&0.001\\
         \midrule
         mIoU&43.4&44.2&\textbf{46.2}&45.4&45.5\\
         \bottomrule
         
    \end{tabular}
    
    \label{weights}
\vspace{-1mm}
\end{table}
\begin{table}[t!]
    \centering
    \caption{Influence of the number of semantic feature samples to be stored ($w$) given $\lambda_{ci}=0.01$}
    \small
    \begin{tabular}{l|ccccc}
        \toprule
         $w$&10&50&200&800&1600\\
         \midrule
         mIoU&45.2&\textbf{46.2}&46.1&45.3&45.0\\
         \bottomrule
    \end{tabular}
    \label{number}
\vspace{-3mm}
\end{table}

\begin{figure*}
\centering
\includegraphics[width=1\textwidth]{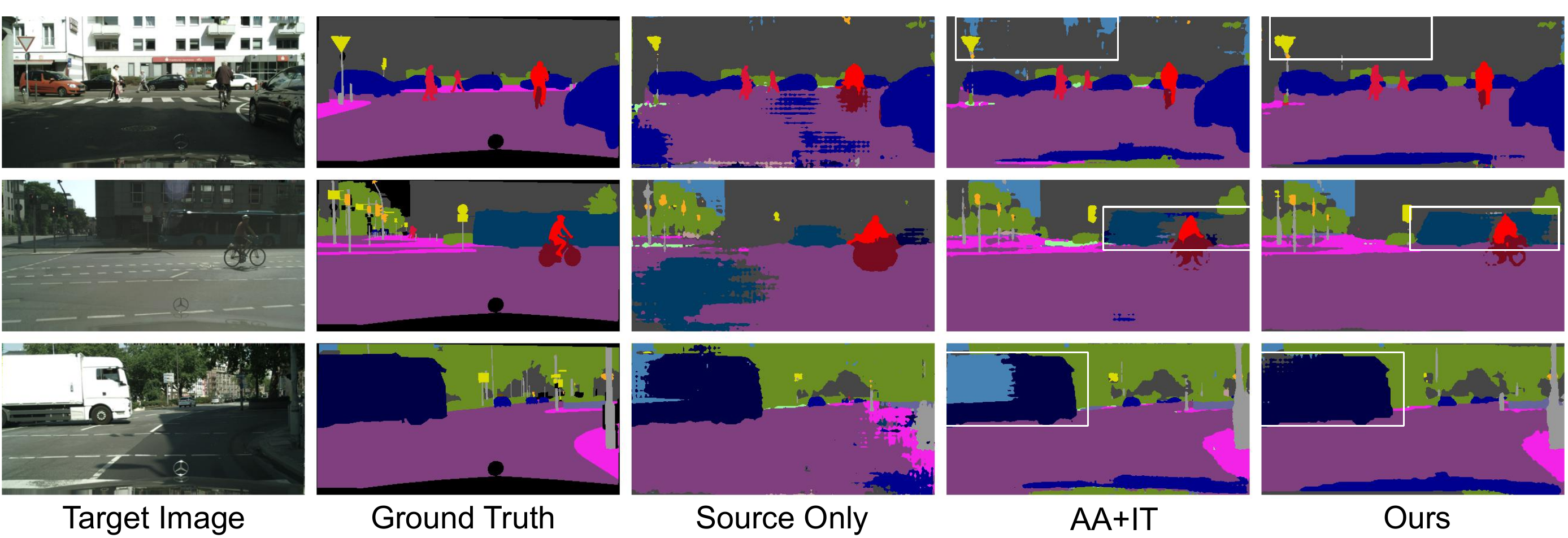}
\caption{Visualization of the segmentation results. 'Source only', 'AA+IT', and 'Ours' correspond to the models that achieves mIoU of 36.6, 44.9, and 49.2 in table \ref{ablation}, respectively. }
\label{output}
\end{figure*}

\begin{table*}[t!]
\caption{Comparison to the state-of-the-art results of adapting SYNTHIA to Cityscapes. }
\small
\begin{center}
\begin{tabular}{ l|p{0.62cm}p{0.62cm}p{0.62cm}p{0.62cm}p{0.62cm}p{0.62cm}p{0.62cm}p{0.62cm}p{0.62cm}p{0.62cm}p{0.62cm}p{0.62cm}p{0.62cm}c }
 \toprule
 \multicolumn{15}{c}{SYNTHIA $\rightarrow$ Cityscapes} \\
 \midrule
Method & \begin{turn}{45}road\end{turn}
& \begin{turn}{45}sidewalk\end{turn}& \begin{turn}{45}building\end{turn}& \begin{turn}{45}light\end{turn}& \begin{turn}{45}sign\end{turn}& \begin{turn}{45}vegetation\end{turn}& \begin{turn}{45}sky\end{turn}& \begin{turn}{45}person\end{turn}& \begin{turn}{45}rider\end{turn}& \begin{turn}{45}car\end{turn}& \begin{turn}{45}bus\end{turn}&  \begin{turn}{45}motorbike\end{turn}& \begin{turn}{45}bike\end{turn}& mIoU
\\
\midrule
Luo et al. \cite{Luo_2019_CVPR} &82.5 &24.0& 79.4& 16.5& 12.7& 79.2& 82.8& 58.3& 18.0& \textbf{79.3}& 25.3& 17.6& 25.9& 46.3 
\\
Tsai et al.\cite{outputspace}&84.3& 42.7& 77.5& 4.7& 7.0& 77.9& 82.5& 54.3& 21.0& 72.3& 32.2& 18.9& 32.3& 46.7
\\
Du et al. \cite{Du_2019_ICCV}& 84.6& 41.7& \textbf{80.8}& 11.5& 14.7& \textbf{80.8}& \textbf{85.3}& 57.5& 21.6& 82.0& 36.0& 19.3& 34.5& 50.0
\\
Li et al. \cite{bidir} &\textbf{86.0}& \textbf{46.7}& 80.3& 14.1& 11.6& 79.2& 81.3& 54.1& 27.9& 73.7& \textbf{42.2}& 25.7& 45.3& 51.4
\\
\midrule
ours (ResNet101) &83.0&44.0&80.3&\textbf{17.1}&\textbf{15.8}&80.5&81.8&\textbf{59.9}&\textbf{33.1}&70.2&37.3&\textbf{28.5}&\textbf{45.8}&\textbf{52.1}
\\
\bottomrule

\end{tabular}
\end{center}
\label{syn2city}
\vspace{-5mm}
\end{table*}

\begin{table}[t!]
    \centering
    \caption{Ablation study on the adaptation from SYNTHIA dataset to Cityscapes dataset. AA stands for adversarial adaptation; IT stands for image transferring; SIM stands for semantic and instance matching; SSL stands for self-supervised learning.}
    \small
    \begin{tabular}{l|cccc|c}
        \toprule
         method & AA & IT & SIM & SSL & mIoU  \\
         \midrule
         source only&&&&&38.6\\
         \,+ AA\cite{outputspace}&\checkmark&&&&45.9\\
         \,+ IT\cite{bidir}&\checkmark&\checkmark&&&46.0\\
         \midrule
         \,+ SIM&\checkmark&\checkmark&\checkmark&&47.1\\
         \,+ SSL&\checkmark&\checkmark&\checkmark&\checkmark&52.1\\
         \midrule
         target only&&&&&71.7\\
         \bottomrule
    \end{tabular}
    \label{synablation}
\vspace{-2mm}
\end{table}

Then, we add our SIM module to the training framework. The background classes include road, sidewalk, building, wall, fence, vegetation, terrain and sky. The foreground classes are all the rest classes used for evaluation. With the best setting for the SIM module where $\lambda_{ci}=0.01$ and $w$, the number of semantic source domain feature samples to be stored, is 50, the mIoU improves to 46.2 by optimizing the Eqn (\ref{Lstep1}). In this setting, we empirically set the maximum source domain instance features of each class to be stored to 10 for each image, and the feature of the instance covering larger area is to be stored with higher priority. We also adapt 10 instance features at maximum for each class from the target domain to the source domain. This is because instance feature representations of small regions or noise regions may be too many for storage and adaptation. 

Finally, we retrain our model with the combination of SIM and the self supervised learning (SSL) framework given the pseudo-labeled target dataset by the training step 1. When generating the pseudo labels for the target dataset, we choose the confidence threshold for each class respectively. We first follow Eqn (\ref{labelpseudo}) to give pseudo labels for each pixel by setting $y_t=0$ for each image in the target dataset. Then, we generate a confidence map corresponding to the pseudo label map where the confidence is the maximum item of the softmax output in each channel so that the pseudo label at each pixel is associated with a confidence value. After this, we rank the confidence values belonging to the same class across the whole target dataset. If the median confidence value is below 0.9, then the confidence threshold for that class is set to the median confidence value; otherwise, it is set to 0.9. With the new $y_t^k$ being set, we follow Eqn (\ref{labelpseudo}) to generate the pseudo labels with don't cares for the target dataset and thus the model retraining can be processed by optimizing Eqn (\ref{Lstep2}). This improves the mIoU to 49.2. 
%
We provide a visualization showing the improvements of our methods in figure \ref{output}.

\textbf{Hyper parameters analysis}. This mainly deals with the settings of $\lambda_{ci}$, the weight for the semantic matching loss and the instance matching loss, and $w$, the number of semantic feature samples to be stored for our proposed SIM module. For the hyper parameters of other modules, we follow \cite{outputspace} to set $\lambda_{seg}=1$, $\lambda_{adv}=0.01$ and $\lambda_D=1$ to control the variables. 

First, we discuss the influence of $\lambda_{ci}$ given $w=50$, which is shown in table \ref{weights}. We experiment the influence of $\lambda_{ci}$ with different $w$'s. Here we only exhibit the results with $w=50$, the setting that achieves the best performance, to provide the intuition of the influence of the choice of $\lambda_{ci}$. We argue that $\lambda_{ci}$ should not be set either too large or too small. If it is too large, the features corresponding to the image-level or instance-level semantic class would be pulled closer to the same source domain feature sample too much, such that these target-domain features would also be very close to each other thus lack intra-class feature variance. This could worsen the scene understanding for the feature extractor and thus negatively impact the overall performance of our model. On the other hand, if $\lambda_{ci}$ is too small, the matching loss would not help the model much on minimizing the feature discrepancy between the source domain and the target domain. As shown in table \ref{weights}, when $\lambda_{ci}=0.01$, an appropriately large value, the model achieves the best performance. 

Second, we show the influence of the choice of $w$, the number of semantic feature samples to be stored, as shown in table \ref{number}. As the model is always being updated during the training stage, it would be infeasible to access all the source-domain feature samples with the newly updated model. Therefore, we store an amount of feature samples generated with recent updated models. The number of these feature samples, $w$, should balance the factors such that 1) $w$ should be large enough so that there will be enough source domain feature samples to be matched; and 2) $w$ should not be too large or the stored source domain feature samples are not up-to-date. With our experiments, $w=50$ achieves the best performance.

\vspace{-2mm}
\subsection{SYNTHIA to Cityscapes}
\vspace{-2mm}
We evaluate the mIoU of 13 classes shared between the source domain and the target domain as \cite{Luo_2019_CVPR, outputspace,Du_2019_ICCV,bidir}.
We use the same hyper parameters which achieves the best performance discussed in section \ref{section52} for all the following experiments.
We compare our model with the previous state-of-the-arts in table \ref{syn2city}. Our model also achieves a new state of the art on adaptation from SYNTHIA dataset to the Cityscapes dataset. 

Table \ref{synablation} shows the contribution of each module.
The model can achieve an mIoU of 38.6 if trained on the source domain only. By adding the adversarial training module and utilizing the transferred source domain images, the model can achieve an mIoU of 46.0. We notice that the improvement of utilizing the transferred images is not obvious, and we conjecture that this is because of the large gap of layouts between the source domain and the target domain. By adding our SIM module, the mIoU improves to 47.1. After retraining our model with self-supervised learning using the same pseudo-labeling strategy described in section \ref{section52}, our model achieves an mIoU of 52.1.


\vspace{-2mm}
\section{Conclusions}
\vspace{-1mm}
We propose a stuff and instance matching (SIM) module for the unsupervised domain adaptation of semantic segmentation from a synthetic dataset to a real-image dataset. We (1) consider the difference of appearance variance between the stuff regions and the instances of things, and thus treat them differently in the adaptation process; (2) explicitly minimize the distance of the closest stuff and instance features between the source domain and the target domain, which enables the adaptation in a more accurate direction and stabilize the GAN training process at longer iterations. By combining our SIM module with self-training, our model achieves a new state-of-the-art on this task. 
\vspace{-2mm}

\paragraph{Acknowledgments}
This work is in part supported by IBM-Illinois Center for Cognitive Computing Systems Research (C3SR) - a research collaboration as part of the IBM AI Horizons Network, and ARC DECRA DE190101315.

{\small
\bibliographystyle{ieee_fullname}
\bibliography{egbib}
}

\end{document}